\documentclass[10pt,twocolumn,letterpaper]{article}

\usepackage{cvpr}
\usepackage{times}
\usepackage{epsfig}
\usepackage{graphicx}
\usepackage{amsmath}
\usepackage{amssymb}

\usepackage[breaklinks=true,bookmarks=false]{hyperref}

\cvprfinalcopy 

\def\cvprPaperID{****} 

\begin{document}

\title{Hierarchical Deep Recurrent Architecture for Video Understanding}

\author{
Luming Tang$^1$ ~~~ Boyang Deng$^2$ ~~~ Haiyu Zhao$^3$ ~~~ Shuai Yi$^3$ \\
$^1$Tsinghua University ~~~ $^2$Beihang University ~~~ $^3$ SenseTime Group Limited \\
{\tt\small {tlm14@mails.tsinghua.edu.cn  ~~~ billydeng@buaa.edu.cn ~~~ \{zhaohaiyu,yishuai\}@sensetime.com}}
}

\maketitle

\begin{abstract}
   This paper \footnote{Source code of the project can be found via the following link \texttt{https://github.com/Tsingularity/youtube-8m}. All the authors share equal contribution in the challenge and preparation of the manuscript.}
   introduces the system we developed for the Youtube-8M Video Understanding Challenge, in which a large-scale benchmark dataset \cite{abu2016youtube} was used for multi-label video classification. 
   The proposed framework contains hierarchical deep architecture, including the frame-level sequence modeling part and the video-level classification part.
   In the frame-level sequence modelling part, we explore a set of methods including Pooling-LSTM (PLSTM), Hierarchical-LSTM (HLSTM), Random-LSTM (RLSTM) in order to address the problem of large amount of frames in a video. We also introduce two attention pooling methods, single attention pooling (ATT) and multiply attention pooling (Multi-ATT) so that we can pay more attention to the informative frames in a video and ignore the useless frames. 
   In the video-level classification part, two methods are proposed to increase the classification performance, i.e. Hierarchical-Mixture-of-Experts (HMoE) and Classifier Chains (CC).
   Our final submission is an ensemble consisting of 18 sub-models. In terms of the official evaluation metric Global Average Precision (GAP) at 20, our best submission achieves \textbf{0.84346} on the public 50\% of test dataset and \textbf{0.84333} on the private 50\% of test data.
\end{abstract}

\section{Introduction}

Video understanding is one of the core tasks in the field of computer vision. The YouTube-8M dataset \cite{abu2016youtube} is a large-scale video understanding dataset consisting of over 7 million YouTube videos which are annotated with 4,716 tags from 25 vertical categories. The average number of labels per video is 3.4.

In the Kaggle competition, Google Cloud \& YouTube-8M Video Understanding Challenge ({\tt https://www.kaggle.com/c/youtube8m}), the dataset is divided into three parts. The training set has 4.9 million videos, the validation set contains 1.4 million videos and the testing set contains 0.7 million samples. 
Moreover, the testing set is divided into two parts, one for public leaderboard evaluation and the other for private leaderboard evaluation.
Participants submit their predictions on the whole test set but could only see the score on public set.
The final score is determined by the score on the private testing set.
In the competition, submissions are evaluated using Global Average Precision (GAP) at 20.
For each video, participants submit a list of predicted labels and their corresponding confidence scores and the evaluation server takes the predicted labels that have the highest 20 confidence scores for each video, then treats each prediction and the confidence score as an individual data point in a long list of global predictions, to compute the Average Precision across all of the predictions and all the videos. 
In detail, the evaluation metric (GAP) is calculated as:

\begin{equation}
GAP=\sum_{i=1}^{N}p(i)\Delta r(i).
\end{equation}

In the rest of our report, we summarize our detailed solution methods in the competition. 
Our baseline models are first introduced and evaluated.
Then our methods on frame-level feature modelling and video-level feature classification are introduced.
Our proposed approach mainly addresses the following three problems.
\begin{itemize}
\item There are too many useless frames in one video for classification. How could we pay more attention to the real informative frames?
\item The large number of frames in a video results in inefficient training. How could we make use of the fact that neighbourhood frames are quite similar?
\item How could we make use of the semantic relation between different labels to increase the performance of classification?
\end{itemize}
At last, we give the ensemble method and models we adopt in our final submission. 
Finally, we summarize our contributions of our submission and propose future work.

\section{Methods}

In this section, we mainly demonstrate all the methods we utilize to get the final submission in details. The whole pipeline could be divided into three parts, frame-level feature modelling, video-level feature classification, and model ensemble.

\subsection{Baseline Models}

In this part, we show the performance of the models proposed in the benchmark paper \cite{abu2016youtube} by using the YouTube-8M Tensorflow Starter Code \footnote{\tt https://github.com/google/youtube-8m}. 
Due to the fact that the dataset used in the Kaggle Competition is a little bit smaller than the dataset reported in the benchmark paper, we also get different results when using the exact same model. 
Here we list results of three models in table \ref{baseline} as baselines, including the Mixture-of-Experts(MoE), Deep Bag-of-Frames (DBOF), Long-Short Term Memory (LSTM).
One model is video level feature based and the other two are frame level feature based. 
We choose a mixture of 8 for the MoE classifier and a base learning rate 0.001 for all these models. 
For the LSTM model, we choose the number of layers to be 1 and the number of cells per layer to be 1024.

\begin{table}
\begin{center}
\begin{tabular}{|l|c|c|}
\hline
Input Feature & Model & GAP \\
\hline\hline
Video Level & Mixture-of-Experts & 0.7930 \\
\hline\hline
Frame Level & Deep-Bag-of-Frames & 0.7948 \\
Frame Level & LSTM & 0.8036 \\
\hline
\end{tabular}
\end{center}
\caption{The GAP of three baseline models, i.e. MoE, DBOF, and LSTM.}
\label{baseline}
\end{table}

\subsection{Frame-level Feature Modelling}

In this part, we describe all the models we propose based on the frame level feature.
On the one hand, the models can be classified as single attention methods and multiple attention methods, and the multiple attention models are trying to pay more attention to informative frames when merging features.
On the other hand, in order to address the time-consuming and hard-to-train problem caused by the large amount of frames in a video, three model types are proposed, such as the MaxPooling-BiLSTM model (MPLSTM), the Hierarchical-BiLSTM model (HLSTM), and the Random-BiLSTM model (RLSTM).
For the convenience of comparison between different models, we use a mixture of 4 for the MoE classifier for all the models in this section.


\subsubsection{Single Attention Based BiLSTM }

\begin{figure}[!h]
\centering
\includegraphics[width=1.0\columnwidth]{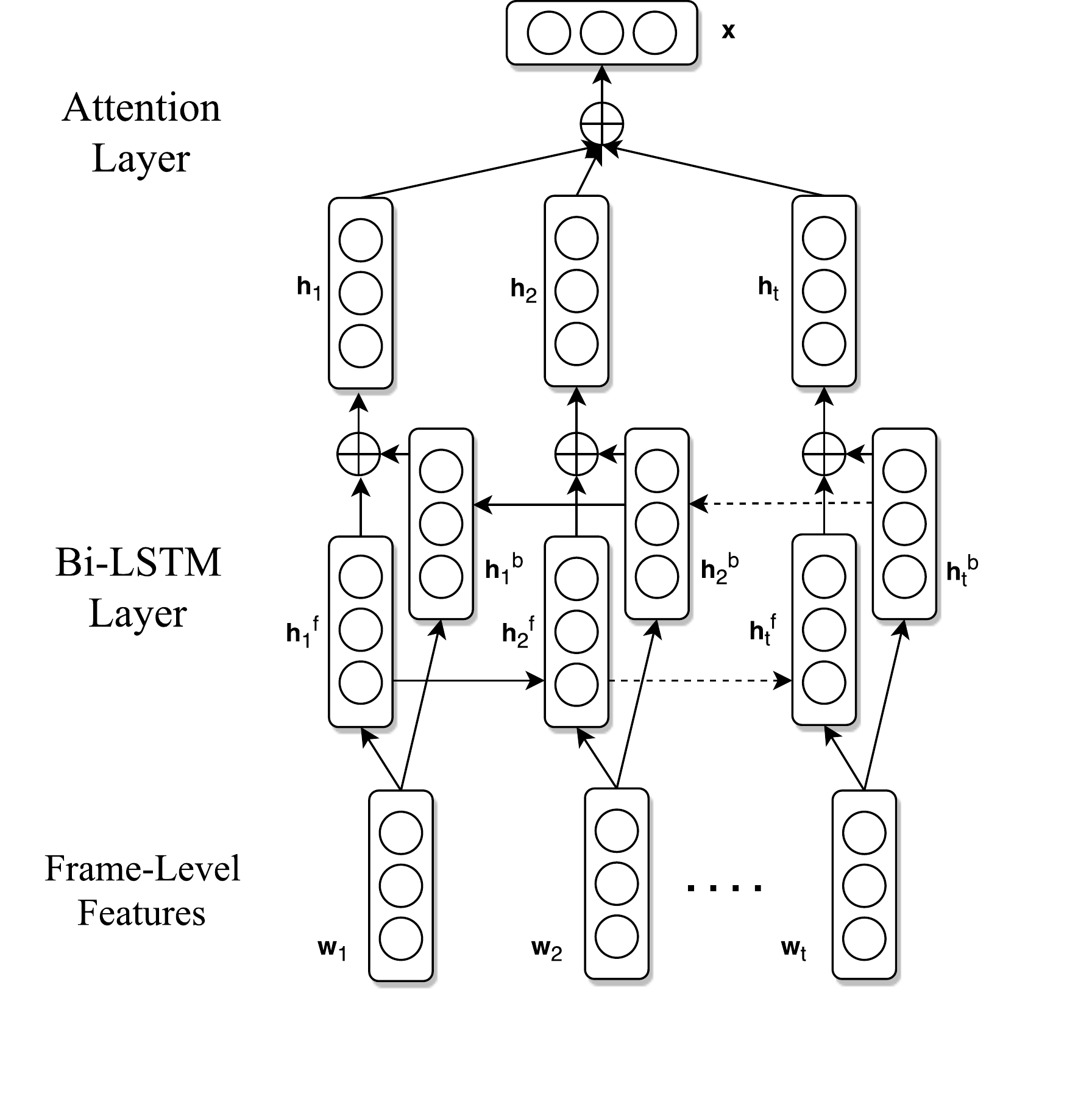}
\caption{The architecture of attention based Bi-LSTM.}
\label{fig:gru}
\end{figure}

This model is proposed to address the problem that there are lots of useless frames in a video frame sequence, e.g. the white/black internal frames. 
It is necessary for us to pay more attention (bigger weights) to the important frames and less attention (smaller weights) to the unimportant frames.

As shown in Fig. \ref{fig:gru}, we encode the single video $x$ into its distributed representation $\mathbf{x}$ by a single attention based Bi-LSTM neural network. 
Firstly, each frames are transformed into its distributed embedding vectors, which we got from the provided tfrecord data. 
Secondly, Bi-LSTM layer is used to extract further features for each frame. 
Thirdly, attention layer is adopted to merge the features into a video-level feature vector as the distributed representation of the video, i.e, $\mathbf{x}$

Long Short-Term Memory (LSTM) units are firstly proposed by \cite{hochreiter1997long} to address the gradient vanishing problem in recurrent neural network by coming up with an adaptive gating mechanism to decide how much the LSTM units keep the previous state as well as memorize the current input data. 
There are lots of variants of LSTM and we adopt a variant called Gated Recurrent Unit (GRU) proposed by \cite{cho2014learning}.
It mainly combines the forget and input gates into a single “update gate” and also merges the cell state and hidden state, which leads to a higher training speed.
In detail, $\mathbf{z}_{t}$ denotes the update gate with corresponding weight matrix $\mathbf{W}^{z}$ and $\mathbf{U}^{z}$, $\mathbf{r}_{t}$ denotes the reset gate with corresponding weight matrix $\mathbf{W}^{r}$ and $\mathbf{U}^{r}$, $\mathbf{\hat{h}}_{t}$ denotes the cell state of step $t$ with corresponding weight matrix $\mathbf{W}^{h}$ and $\mathbf{U}^{h}$ and $\mathbf{h}_{t}$ denotes the output value of step $t$. 
The whole process in Fig.\ref{fig:gru} could be formulated as follows,
\begin{equation}
\mathbf{z}_{t}=\sigma(\mathbf{W}^z \mathbf{w}_{t}+\mathbf{U}^z \mathbf{h}_{t-1}),
\end{equation}
\begin{equation}
\mathbf{r}_{t}=\sigma(\mathbf{W}^r \mathbf{w}_{t}+\mathbf{U}^r \mathbf{h}_{t-1}),
\end{equation}
\begin{equation}
\mathbf{\hat{h}}_{t}=\tanh(\mathbf{W}^{h} \mathbf{w}_{t}+\mathbf{U}^{h}(\mathbf{h}_{t-1}\odot \mathbf{r}_{t})),
\end{equation}
\begin{equation}
\mathbf{h}_{t}=(1-\mathbf{z}_{t})\odot \mathbf{\hat{h}}_{t}+\mathbf{z}_{t}\odot \mathbf{h}_{t-1},
\end{equation}
where $\sigma$ denotes the sigmoid function and $\odot$ denotes the element-wise multiplication. 
The initial hidden state $\mathbf{h}_{0}$ is fixed to $\mathbf{0}$.

Due to the fact that the future context information is also as important as past information, we use the bidirectional LSTM networks which contain both forward and backward temporal hidden state connection order and make the model be able to exploit information from both the past and the future. Therefore, the network consists of two sub-networks for the left and right sequence context and then the output of the i-th frame $\mathbf{w}_{i}$ is $\mathbf{h}_{i}^{f}$ and $\mathbf{h}_{i}^{b}$, representing forward and backward respectively. Here, we use element-wise sum to merge them together to get each frame's feature vector $\mathbf{h}_{i}$:
\begin{equation}
\mathbf{h}_{i}=\mathbf{h}_{i}^{f} \oplus \mathbf{h}_{i}^{b}.
\end{equation}

Due to the fact that different frames in one video contain different amount of useful information for video category classification, we should pay each frame different level of attention instead of simple averaging when merging features. 
Attention mechanism is firstly introduced by \cite{bahdanau2014neural} in order to stress the target words step by step in machine translation and here we use this method to help extract a better feature vector for single video. 
The single attention method we use here is quite similar to the method proposed by \cite{zhou2016attention}, in which attention is used for relation extraction, a natural language processing (NLP) task. 
Suppose $\mathbf{H} = [\mathbf{h}_{1}, \mathbf{h}_{2}, \cdots, \mathbf{h}_{T}]$ is the matrix consisting of all output vectors that produced by the Bi-LSTM layer, the whole video's feature vector $\mathbf{x}$ is formed by a weighted sum of each step's output:
\begin{equation}
\mathbf{M}=\tanh(\mathbf{H}),
\end{equation}
\begin{equation}
\mathbf{\alpha}=Softmax(\mathbf{\omega}^{T}\mathbf{M}),
\end{equation}
\begin{equation}
\mathbf{r}=\mathbf{H}\mathbf{\alpha}^{T},
\end{equation}
\begin{equation}
\mathbf{x}=\tanh(\mathbf{r}),
\end{equation}
where  $\mathbf{H} \in \mathbb{R}^{d^h*T}$, $d^h$ is the dimension of hidden state in GRU network and $T$ is the amount of frames in one video, $\mathbf{\omega}$ is a trainable query vector and $\omega^{T}$ indicates its transposition.

Before fed into the MoE classifier, we also adopt dropout \cite{srivastava2014dropout} on the video-level feature vector $\mathbf{x}$ to alleviate overfitting. The dropout layer is defined as follows:
\begin{equation}
\mathbf{x}=\frac{1}{p}(\mathbf{x}\circ \mathbf{h}),
\end{equation}
where $\circ$ denotes an element-wise multiplication and $\mathbf{h}$ represents a vector of Bernoulli random variables with probability $p$. $\frac{1}{p}$ indicates the feature scaling while training. In the inference phase, $p=1$.

We also adopt the same single attention method, Bi-LSTM network and dropout method in the following models. 

Besides, we also tried splitting method, in which visual features (1024 dimension) and audio features (128 dimension) are fed into the ATT+BiLSTM network individually to get each video-level feature vector, $\mathbf{x}_{visual}$, $\mathbf{x}_{audio}$ and concatenated to get the final feature vector at last, i.e., $\mathbf{x}=\mathbf{x}_{visual} \oplus \mathbf{x}_{audio}$, $\oplus$ stands for concatenate operation here.

\subsubsection{Multiply Attention Based BiLSTM }

In the previous part, we introduce the single attention method which greatly increase the classification performance. However, it ignore the fact that when predicting different labels, we should pay attention to different frames of the video. For instance, when predicting the label "football", we should put more weights on the frames containing green grass and football players, and when facing the label "cooking", we should pay more attention to the frames that have red meat and green vegetables. As a result, different labels should have different attention query vectors instead of the same single vector.

However considering the number of labels is 4716, which is so large that it is easy to be overfitting as well as time consuming if we truly use 4716 attention query vectors. Due to the fact that labels that share similar semantics could also share attention vectors, so we use 25 attention vectors instead, in accordance with the 25 vertical labels. As a result, a video should have 25 video-level representation feature vectors, i.e, $\{\mathbf{x}_1,\mathbf{x}_2,\cdots,\mathbf{x}_{25}\}$. For instance, labels are $\{a_1,a_2,\cdots,a_{4716}\}$ and vertical labels are $\{b_1,b_2,\cdots,b_{25}\}$ when predicting the score of label $a_i$ and the label $a_i$ belongs to the vertical label $b_j$, we should use the feature vector $\mathbf{x}_j$ to represent the whole video and do the following classification.

In detail, for each video-level feature vector $\mathbf{x}_i$ is computed as a weighted sum of each frame's post Bi-LSTM feature $\mathbf{h}_i$:
\begin{equation}
\mathbf{x}_i=\sum_{j} \alpha_{i,j}\mathbf{h}_{j},
\end{equation}
where $\alpha_{i,j}$ is the weight of each frame vector and here we use a selective attention proposed by \cite{lin2016neural} to calculate the weight.
So $\alpha_{i,j}$ is defined as:
\begin{equation}
\alpha_{i,j} = \frac{\exp( e_{i,j} )}{\sum_k\exp( e_k)},
\end{equation}
where $e_{i,j}$ is a query-based function which scores how well the current frame $h_{j}$ and the target vertical label $b_{i}$ matches:
\begin{equation}
e_{i,j} = \mathbf{h}_{j}\mathbf{A}\mathbf{b}_{i},
\end{equation}
where $\mathbf{A}$ is a weighted diagonal matrix and query vector $\mathbf{b}_{i}$ is the representation vector of vertical label $b_{i}$.

\begin{table}
\begin{center}
\begin{tabular}{|l|c|}
\hline
Model & GAP \\
\hline\hline
LSTM & 0.80357 \\
BiLSTM & 0.80534 \\
\hline\hline
Single-ATT+BiLSTM & \textbf{0.81308} \\
Split+Single-ATT+BiLSTM & \textbf{0.81309} \\
\hline\hline
Multi-ATT+BiLSTM & 0.81187 \\
\hline
\end{tabular}
\end{center}
\caption{The performance of attention based Bi-LSTM.}
\label{attention}
\end{table}

The performance of each attention based model is listed in the table \ref{attention}. It is quite obvious that both bidirectional network and attention method make big contribution to the model's gain on GAP, especially the single attention method. In contrast, the splitting method is quite time consuming but couldn't get obviously better performance, as a result, we won't apply splitting operation in the following models. However, in practice splitting models could work well during ensemble thus we use quite a lot splitting models in the final submission. As for the multiply attention method, it is obvious that although it also increase the performance of Bi-LSMT, its performance is still worse than single attention models, which is quite confusing. We guess maybe more intermediate supervision could make it better, such as the loss function on the vertical label classification. In summary, the single attention method works the best and most efficiently, so we adopt this attention method in the models introduced in the following section.  

\subsubsection{Other Single Attention Based BiLSTM Models}

In this part, we will introduce three models based on the previous single attention BiLSTM model, MaxPooling, Random, and Hierarchical. They are all proposed to address the issues caused by too many frames in a video, which leads to gradient vanishing and hard-to-train problem for recurrent neural network. 
In our dataset, most videos are over 150 frames per second which is indeed too long for LSTM to reach its best performance. Besides, simply going over every frame feature in recurrent network is also a waste of time because the neighbourhood features are usually quite similar because of the video sequence's consistency.

\begin{figure}[!h]
\centering
\includegraphics[width=1.0\columnwidth]{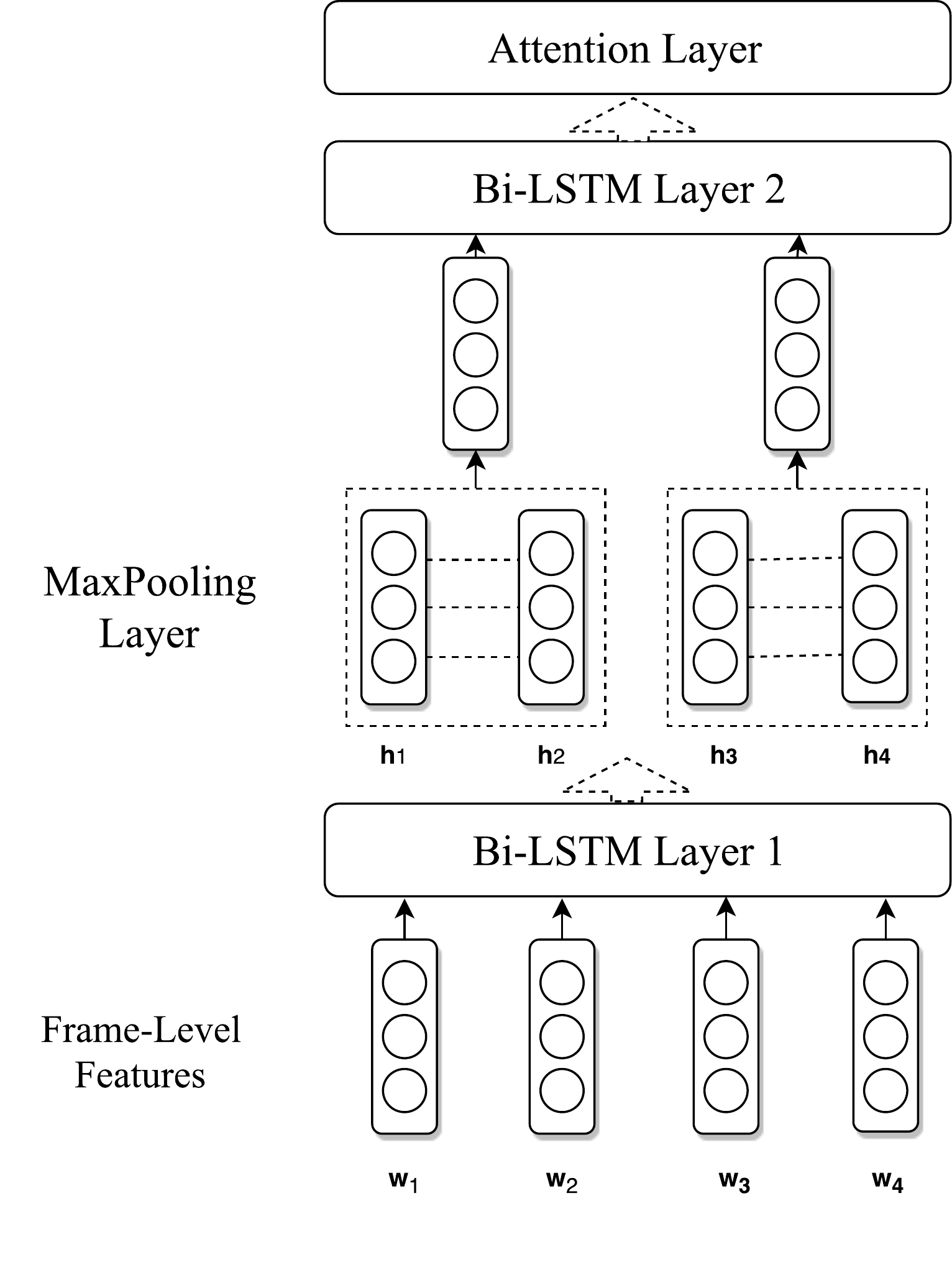}
\caption{The architecture of attention based MaxPooling Bi-LSTM.}
\label{fig:pool}
\end{figure}

The first model is Single Attention Based MaxPooling BiLSTM, which could be illustrated in Fig.\ref{fig:pool}. It just simply inserts max-pooling operation between two BiLSTM layers to merge neighbour frames' feature as well as reduce the number of frames for the second BiLSTM. The window size of max-pooling is a hyper-parameter and after several experiments we select 3 and 5.

The second model is Single Attention Based Random BiLSTM, which could be viewed as a data augmentation method. It randomly samples one frame in each five neighbour frames thus the input of the whole neural network is $max\_frames/5=300/5=60$ frames. At such length, it is much easier and more efficient to learn a deeper LSTM so we can change the previous hyperparameter "LSTM layers" from 1 to 4 or even larger to increase the performance. Besides, because the input is always changing even though it is the same video, the network contains stronger robustness.

\begin{figure}[!h]
\centering
\includegraphics[width=1.0\columnwidth]{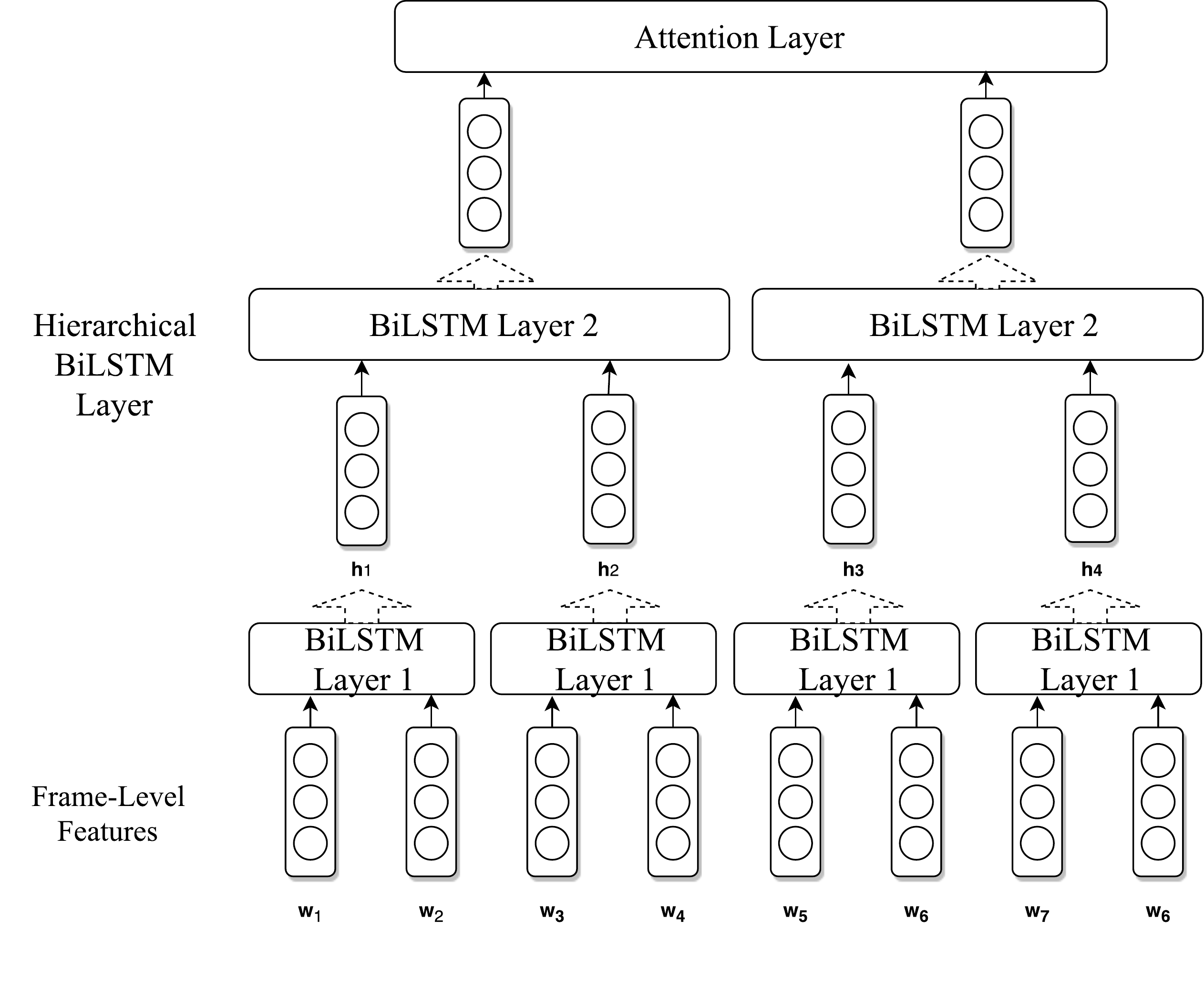}
\caption{The architecture of attention based Hierarchical Bi-LSTM.}
\label{fig:hlstm}
\end{figure}

The third model is Single Attention Based Hierarchical BiLSTM, whose idea is quite similar to the max-pooling one. The only difference is that, instead of using max-pooling operation, it uses smaller BiLSTM to merge neighbourhood frame features. The whole network could be shown in Fig. \ref{fig:hlstm}. Each BiLSTM layer share the same LSTM cell parameters.

The performance of these models are listed in table \ref{pool}. It is obvious that max-pooling method gets best performance and window\_size 3 is the best choice. Through both the random sampling and deeper LSTM network, the Random method also gets better performance than the baseline attention based BiLSTM model. However, Hierarchical method even reduces the GAP which is quite confusing, maybe because of its much more huge amount of parameters. But it still counts when ensemble.

\begin{table}
\begin{center}
\begin{tabular}{|l|c|}
\hline
Model & GAP \\
\hline\hline
ATT+BiLSTM & 0.81308 \\
\hline\hline
ATT+MaxPool+BiLSTM (window\_size=3) & \textbf{0.82046} \\
ATT+MaxPool+BiLSTM (window\_size=5) & 0.81821 \\
\hline\hline
ATT+Random+BiLSTM (layer=1) & 0.80954 \\
ATT+Random+BiLSTM (layer=4) & \textbf{0.81513} \\
\hline\hline
ATT+Hierarchical+BiLSTM & 0.80934 \\
\hline
\end{tabular}
\end{center}
\caption{The performance of each single attention based BiLSTM.}
\label{pool}
\end{table} 

\subsection{Video-level Feature Classification}

The network structures elaborated above have showed extraordinary power to abstracting
critical features from sequential data. Based on these achievements, we found that we have
gracefully obtained a set of updated video level features, \ie the aggregated features
produced by our frame level models. More specifically, we dig more potential of data from
the perspective of correlated classes. As showed in the final results, our proposed video
level models, \textbf{HMoE} and \textbf {CC}, achieve outstanding single-model
performances based on aggregated features. 

\subsubsection{Hierarchical Mixture of Experts (HMoE)}

A great obstacle of this task is that we have a huge number of classes to recognize.
Common sense tells us that the more classes we need to distinguish, the more efforts we
need to put into learning. Granted, it is the case for absolutely independent classes.
However, when it comes to correlated classes, we may find that more classes can sometimes
bring us more information to understand the videos or images. Hence, we first investigate
the cluster property of $4716$ target classes, which leads to a Hierarchical Mixture of
Experts model.

Naturally, there are some coarse classes offered together with those actual fine classes.
These labels are precious resources to be harnessed. In order to make use of this original
clustering information, we adjust the simple one-layer MoE model into a two level
hierarchical model. In fact, each of these levels is a MoE model for corresponding
classes. The difference is that the higher level classifier for fine classes are
conditional probabilities given the probabilities of coarse classes.

On implementation, we concatenate the responses before softmax layer from the classifier
for coarse classes with
aggregated features and feed them into the classifier for fine classes. During training,
we employ the given coarse labels to supervise the coarse classifier. The two levels of HMoE
is trained jointly. Considering the fact that the number of coarse classes is incomparable
to the number of fine classes, we can obtain a more reasonable hierarchical model without
hurting the efficiency of the original MoE model. Especially, our best single model, which
achieved \textbf{$82.416$} GAP, deployed an HMoE on its top.

\subsubsection{Classifier Chains (CC)}

Apart from the clustering property, we know there can be correlations among different
classes. Some of them are conflicts, \eg a cat can't be a dog at the same time. Some of
them are directed boundings, \eg a basketball must be a ball. We believe that these
relationships can play a vital role in classification tasks, especially in multi-label
classification. Previous literature has tried to address this issue. Reed \etal
\cite{read2011classifier}
presented a chaining method to model inter-dependencies between models. Recently, Wang
\etal \cite{wang2016cnn} proposed an improved method using recurrent neural networks to exploit
correlations between labels in an elegant way. Different from their methods, we adapt the
spirit of Classifier Chains to this specific challenge. As a result, we developed a CC for
YouTube-8M.

A major difference of this challenge from previous tasks is that we have more than $4000$
classes to distinguish. That is to say, we can't afford making binary classifications for
each class sequentially. On tackling this obstacle, we invent a group strategy, which is a
trade-off between efficiency and effectiveness. In other words, we ordered these classes on
their frequency in the training set. Then, we sequentially make a group every $393$
classes. The divided set has $12$ groups with $393$ classes each. Obviously, $12$ groups
are affordable for Classifier Chains. Moreover, in the light of the excellent performances
of MoE and HMoE models, we deploy MoE models as the classifiers in the chain. As for the
correlation among groups, we maintain a whole classes probability distribution in each
run. The input of the classifier for each step is the combination of aggregated features
and $4716$ classes probabilities. Again, it's not affordable for an MoE to handle inputs
in more than $4000$ dimensions. Hence, we borrow the bottle-neck idea from
\cite{he2016deep} to
reduce the dimensions. The final CC is a neat model for multi-label classification. The
best single model with CC on its top achieves $0.823$ GAP on the testing set. We believe
that CC together with HMoE plays an essential role in our final ensemble.

In summary, all the performances of video-level classification are listed in the Table \ref{video}.

\begin{table}
\begin{center}
\begin{tabular}{|l|c|}
\hline
Model & GAP \\
\hline\hline
ATT+MaxPool+BiLSTM & 0.82046 \\
\hline\hline
ATT+MaxPool+BiLSTM+HMoE & \textbf{0.82416} \\
ATT+MaxPool+BiLSTM+CC & 0.82301 \\
\hline
\end{tabular}
\end{center}
\caption{The performance of video-level classification.}
\label{video}
\end{table} 

\section{Model Ensemble}

Our final submission is the average of the following 18 sub models. In fact, we also try some other ensemble methods such as weighted sum, xgboost or even training the weights on validation set, but it turns out that average is still the best. It also shows the strong generalization ability of our models.

If not otherwise specified, the learning rate strategy is the default settings in the starter code. All the models' base learning rate=0.001

1. ATT + MaxPooling + BiLSTM, batch size is 64, number of GPUs is 8, mixture of MoE is 4, cell of LSTM is 1152, layer of LSTM is 1, window size is 3. GAP=0.81613

2. ATT + MaxPooling + BiLSTM, batch size is 128, number of GPUs is 4, mixture of MoE is 8, cell of LSTM is 1152, layer of LSTM is 1, window size is 3. GAP=0.81768

3. ATT + Splitting + BiLSTM, batch size is 128, number of GPUs is 2, mixture of MoE is 16, cell of LSTM is 1152, layer of LSTM is 1. GAP=0.81827

4. ATT + MaxPooling + BiLSTM, batch size is 128, number of GPUs is 2, mixture of MoE is 8, cell of LSTM is 1152, layer of LSTM is 1, window size is 5. GAP=0.81619

5. ATT + Splitting + BiLSTM, batch size is 128, number of GPUs is 1, mixture of MoE is 16, cell of LSTM is 1152, layer of LSTM is 1. GAP=0.81909

6. ATT + Splitting + BiLSTM, batch size is 64, number of GPUs is 2, mixture of MoE is 16, cell of LSTM is 1152, layer of LSTM is 1. GAP=0.81703

7. ATT + MaxPooling + BiLSTM + HMoE, batch size is 64, number of GPUs is 8, mixture of MoE is 4, cell of LSTM is 1152, layer of LSTM is 1, window size is 3. GAP=0.81811

8. ATT + MaxPooling + BiLSTM + Dropout + HMoE, batch size is 64, number of GPUs is 8, mixture of MoE is 4, cell of LSTM is 1152, layer of LSTM is 1, window size is 3, checkpoint is 46441. GAP=0.82416

9. ATT + MaxPooling + BiLSTM, batch size is 128, number of GPUs is 4, mixture of MoE is 8, cell of LSTM is 1152, layer of LSTM is 1, window size is 5. GAP=0.81748

10. ATT + MaxPooling + BiLSTM, batch size is 128, number of GPUs is 4, mixture of MoE is 16, cell of LSTM is 1152, layer of LSTM is 1, window size is 3. GAP=0.81704

11. ATT + MaxPooling + BiLSTM, batch size is 128, number of GPUs is 4, mixture of MoE is 8, cell of LSTM is 1152, layer of LSTM is 1, window size is 3, learning rate decay is 0.5, learning rate decay examples is 2000w. GAP=0.81704

12. ATT + MaxPooling + BiLSTM + Dropout, batch size is 128, number of GPUs is 4, mixture of MoE is 8, cell of LSTM is 1152, layer of LSTM is 1, window size is 3. GAP=0.81865

13. ATT + MaxPooling + BiLSTM + Dropout, batch size is 128, number of GPUs is 4, mixture of MoE is 8, cell of LSTM is 1152, layer of LSTM is 1, window size is 3, learning rate decay is 0.5, learning rate decay examples is 2000w. GAP=0.81996

14. ATT + MaxPooling + BiLSTM + Validation data, batch size is 128, number of GPUs is 4, mixture of MoE is 8, cell of LSTM is 1152, layer of LSTM is 1, window size is 3. GAP=0.81878

15. ATT + Splitting + BiLSTM + Validation data, batch size is 128, number of GPUs is 4, mixture of MoE is 16, cell of LSTM is 1152, layer of LSTM is 1, window size is 3. GAP=0.81885

16. ATT + Splitting + BiLSTM + Dropout, batch size is 128, number of GPUs is 1, mixture of MoE is 16, cell of LSTM is 1152, layer of LSTM is 1, learning rate decay is 0.5, learning rate decay examples is 2000w. GAP=0.81796

17. ATT + MaxPooling + BiLSTM + Dropout + CC, batch size is 64, number of GPUs is 8, mixture of MoE is 4, cell of LSTM is 1152, layer of LSTM is 1, window size is 3. \textbf{GAP=0.82301}

18. ATT + MaxPooling + BiLSTM + Dropout + HMoE, batch size is 64, number of GPUs is 8, mixture of MoE is 4, cell of LSTM is 1152, layer of LSTM is 1, window size is 3, checkpoint is 53725. \textbf{GAP=0.82416}

\section{Summary}

In this report, we describe our methods in the YouTube-8M competition. 
We propose attention method to make the model pay more attention to the informative frames instead of useless frames when merging features. 
We also introduce some feature merging methods to reduce the large number of frames while increasing the performance at the meantime. 
We also propose video-level feature classification method by making use of semantic relation among labels to get much higher GAP. 
However, due to the time limitation, there are still lots of problem we haven't addressed.
For instance, the current attention and label correlation methods are really simple and sometimes naive. 
The processing of video frame sequences is still inefficient compared to human self because the network is always processing each frame however the way of human does, in fact, is more like a jump-reading. 
There are still a lot of work ahead of us in the challenge of video understanding.

{\small
\bibliographystyle{ieee}
\bibliography{egbib}

\begin{thebibliography}{10}\itemsep=-1pt

\bibitem{abu2016youtube}
S.~Abu-El-Haija, N.~Kothari, J.~Lee, P.~Natsev, G.~Toderici, B.~Varadarajan,
  and S.~Vijayanarasimhan.
\newblock Youtube-8m: A large-scale video classification benchmark.
\newblock {\em arXiv preprint arXiv:1609.08675}, 2016.

\bibitem{bahdanau2014neural}
D.~Bahdanau, K.~Cho, and Y.~Bengio.
\newblock Neural machine translation by jointly learning to align and
  translate.
\newblock {\em arXiv preprint arXiv:1409.0473}, 2014.

\bibitem{cho2014learning}
K.~Cho, B.~Van~Merri{\"e}nboer, C.~Gulcehre, D.~Bahdanau, F.~Bougares,
  H.~Schwenk, and Y.~Bengio.
\newblock Learning phrase representations using rnn encoder-decoder for
  statistical machine translation.
\newblock {\em arXiv preprint arXiv:1406.1078}, 2014.

\bibitem{he2016deep}
K.~He, X.~Zhang, S.~Ren, and J.~Sun.
\newblock Deep residual learning for image recognition.
\newblock In {\em Proceedings of the IEEE conference on computer vision and
  pattern recognition}, pages 770--778, 2016.

\bibitem{hochreiter1997long}
S.~Hochreiter and J.~Schmidhuber.
\newblock Long short-term memory.
\newblock {\em Neural computation}, 9(8):1735--1780, 1997.

\bibitem{lin2016neural}
Y.~Lin, S.~Shen, Z.~Liu, H.~Luan, and M.~Sun.
\newblock Neural relation extraction with selective attention over instances.
\newblock In {\em Proceedings of ACL}, volume~1, pages 2124--2133, 2016.

\bibitem{read2011classifier}
J.~Read, B.~Pfahringer, G.~Holmes, and E.~Frank.
\newblock Classifier chains for multi-label classification.
\newblock {\em Machine learning}, 85(3):333--359, 2011.

\bibitem{srivastava2014dropout}
N.~Srivastava, G.~E. Hinton, A.~Krizhevsky, I.~Sutskever, and R.~Salakhutdinov.
\newblock Dropout: a simple way to prevent neural networks from overfitting.
\newblock {\em Journal of Machine Learning Research}, 15(1):1929--1958, 2014.

\bibitem{wang2016cnn}
J.~Wang, Y.~Yang, J.~Mao, Z.~Huang, C.~Huang, and W.~Xu.
\newblock Cnn-rnn: A unified framework for multi-label image classification.
\newblock In {\em Proceedings of the IEEE Conference on Computer Vision and
  Pattern Recognition}, pages 2285--2294, 2016.

\bibitem{zhou2016attention}
P.~Zhou, W.~Shi, J.~Tian, Z.~Qi, B.~Li, H.~Hao, and B.~Xu.
\newblock Attention-based bidirectional long short-term memory networks for
  relation classification.
\newblock In {\em Proceedings of ACL}, page 207, 2016.

\end{thebibliography}
}

\end{document}